\documentclass{IEEEtran4PSCC}
\ifCLASSINFOpdf
   \usepackage[pdftex]{graphicx}
\else
   \usepackage[dvips]{graphicx}
\fi
\usepackage[cmex10]{amsmath}
\usepackage{amsfonts}
\usepackage{amsthm}
\usepackage{amssymb}
\usepackage{bm}
\usepackage{algorithm}
\usepackage[noend]{algpseudocode}
\usepackage{cite}
\usepackage{comment}
\usepackage{xcolor}
\usepackage{multirow}
\usepackage{tabularx}
\newcolumntype{C}{>{\Centering\hspace{0pt}}X}
\makeatletter
\let\old@ps@headings\ps@headings
\let\old@ps@IEEEtitlepagestyle\ps@IEEEtitlepagestyle
\def\psccfooter#1{%
    \def\ps@headings{%
        \old@ps@headings%
        \def\@oddfoot{\strut\hfill#1\hfill\strut}%
        \def\@evenfoot{\strut\hfill#1\hfill\strut}%
    }%
    \def\ps@IEEEtitlepagestyle{%
        \old@ps@IEEEtitlepagestyle%
        \def\@oddfoot{\strut\hfill#1\hfill\strut}%
        \def\@evenfoot{\strut\hfill#1\hfill\strut}%
    }%
    \ps@headings%
}
\makeatother

\newtheorem{lemma}{Lemma}

\renewcommand{\baselinestretch}{0.98}

\begin{document}

%
\title{Modeling the AC Power Flow Equations with Optimally Compact Neural Networks:\\ Application to Unit Commitment}



\author{Alyssa~Kody$^\dagger$,
        Samuel~Chevalier$^\dagger$,
        Spyros~Chatzivasileiadis,
        Daniel~Molzahn}
\maketitle

\begin{abstract}
Nonlinear power flow constraints render a variety of power system optimization problems computationally intractable. Emerging research shows, however, that the nonlinear AC power flow equations can be successfully modeled using Neural Networks (NNs). These NNs can be exactly transformed into Mixed Integer Linear Programs (MILPs) and embedded inside challenging optimization problems, thus replacing nonlinearities that are intractable for many applications with tractable piecewise linear approximations. Such approaches, though, suffer from an explosion of the number of binary variables needed to represent the NN. Accordingly, this paper develops a technique for training an ``optimally compact" NN, i.e., one that can represent the power flow equations with a sufficiently high degree of accuracy while still maintaining a tractable number of binary variables. We show that the resulting NN model is more expressive than both the DC and linearized power flow approximations when embedded inside of a challenging optimization problem (i.e., the AC unit commitment problem).
\end{abstract}

\begin{IEEEkeywords}
AC power flow, AC unit commitment (AC-UC), mixed-integer linear program (MILP), neural networks, piecewise linear model
\end{IEEEkeywords}

\thanksto{\indent $^\dagger$ denotes an equal contribution among authors.\\
\indent This material is based upon work supported by Laboratory Directed Research and Development (LDRD) funding from Argonne National Laboratory, provided by the Director, Office of Science, of the U.S. Department of Energy under Contract No. DEAC02-06CH11357.\\
\indent Alyssa Kody is with Argonne National Laboratory. Email: akody@anl.gov.\\
\indent Samuel Chevalier and Spyros Chatzivasileiadis are with the Center for Electric Power and Energy, Department of Electrical Engineering, Technical University of Denmark (DTU). Emails: \{spchatz; schev;\}@elektro.dtu.dk.\\
\indent Daniel Molzahn is with the  School of Electrical and Computer Engineering, Georgia Institute of Technology. Email: molzahn@gatech.edu.\\
}

\section{Introduction}
The AC power flow equations are routinely used to model network constraints in optimization problems related to the control, operation, and planning of power systems. These constraints, however, are both nonlinear and non-convex, resulting in optimization problems that can be NP-hard~\cite{bienstock2015strong}. 
In practice, many power systems operation problems, like unit commitment (UC), optimal power flow (OPF), and optimal transmission switching (OTS), are solved using linearized approximations of the nonlinear AC power flow equations for the sake of computational tractability.
However, linear approximations can result in suboptimal solutions or solutions that are infeasible in the original, nonlinear problem~\cite{Baker:2021}.

\textit{Piecewise linear models} offer a method of improving solution accuracy by capturing some of the nonlinearity of the AC power flow equations \cite{foster_dissertation, Nanou:2021}. 
One way to construct such a piecewise linear approximation is through the use of Neural Networks (NNs) with rectified linear unit (ReLU) activation functions. 
Recent literature, e.g.,~\cite{ donon2020neural,Hu:2021}, 
has shown that NNs can be used to model the AC power flow equations with a very high degree of accuracy.
Furthermore, NNs using ReLU activation functions can be \textit{exactly} transformed into a Mixed Integer Linear Program (MILP); this can be accomplished by representing the activation of each ReLU function as a binary variable and then using the big-M method to formulate activation constraints as a function of the NN weights and biases \cite{tjeng2017evaluating}.

Although the exact MILP reformulation of a NN is often used for verification purposes \cite{Venzke:2020, venzke2020learning}, it can also be embedded within a larger optimization problem as a function approximation.
This technique allows us to transform a problem that may have originally been a challenging mixed-integer nonlinear program (MINLP) into a more tractable MILP.
Researchers have embedded NNs as MILPs within optimization problems for a variety of applications: \cite{Katz:2020b, Huchette:2020, Say:2017, Grimstad:2019}. 
Specifically in the field of power systems, \cite{Zhang:2020} and \cite{Zhang:2021} encode frequency constraints using the MILP reformulation in microgrid scheduling and UC problems, respectively, and~\cite{Venzke:2020} encodes security constraints into the OPF problem.

In this paper, we develop a NN-based piecewise linear model of AC power flow equations that is both more accurate than a standard linearization and more computationally tractable than the original nonlinear equations.
The resulting NN can be embedded into any optimization problem as a surrogate power flow model via an exact MILP reformulation.
However, the number of binary variables in the MILP reformulation \textit{scales linearly} with the number of hidden neurons.
Therefore, embedding a NN containing many neurons, which may be necessary to represent  high degrees of nonlinearity, can be a computational bottleneck.
Grimstad and Andersson in \cite{Grimstad:2019} observe that ``the feasibility of using the MILP formulation quickly fades with increasing network sizes.'' 

This motivates the use of compression techniques to reduce the number of neurons in a NN, and consequently, the number of binary variables in the MILP reformulation.
The machine learning community has developed a variety of methods to compress NNs with minimal accuracy compromise \cite{Cheng:2018}. 
This paper specifically employs several methods in order to achieve an ``optimally compact" NN model: (1) \textit{low-rank updates}: we exploit the observation that, in most regions of practical interest, the power flow equations exhibit a relatively low degree of nonlinearity, and we therefore learn low-rank updates of a physics-based linearization of the AC power flow equations in these regions, (2) \textit{pruning}: ReLU activation functions that are found to be always active or inactive over the input domain are fixed or removed, respectively, and (3) \textit{sparsification}: we set NN weights below some threshold to be zero and re-train.

We demonstrate the use of the compact AC power flow NN by embedding it as a MILP within the day-ahead UC problem, which aims to determine the minimum cost generator commitment and dispatch while meeting load demand and abiding by physical laws and feasibility constraints. 
The AC unit commitment (AC-UC) problem, which constrains power injection and line flow quantities via the AC power flow equations, is a challenging nonconvex MINLP that is NP-hard \cite{Tseng:1996}.
In practice, 
utilities simplify the
AC-UC problem to the DC unit commitment (DC-UC) problem, thus neglecting reactive power dispatch, line losses, and voltage magnitude constraints.
Hence, corrective actions are often needed to account for these omissions~\cite{liu2018global}. Furthermore, as demonstrated in this paper, a commitment schedule based on a DC-UC solution can result in inoperable AC power dispatching solutions. 

Researchers have proposed various methods to increase the computational tractability of the AC-UC problem using, for example, Lagrange relaxation~\cite{murillo-sanchez1998}, decomposition~\cite{castillo2016, liu2018global}, 
and convexification methods \cite{zohrizadeh2018}. 
However, this is an ongoing research area without a singular superior solution technique identified yet. 
Recently, in \cite{Nanou:2021}, Nanou et al. develop a piecewise linear AC power flow model, which is embedded within a UC problem as a MILP, but their work does not use NNs or learning to generate the piecewise linear model.
Although there are some works that focus on using NNs to solve the power flow equations \cite{donon2019graph, donon2020neural,Hu:2021}, the authors are not aware of existing literature that embeds NN models of the AC power flow equations reformulated as MILPs within the UC problem, nor any other other power systems optimization problem. Accordingly, the contributions of this paper follow:
\begin{enumerate}
    \item Using a sequence of feasible power flow solutions, we learn a piecewise linear power flow mapping based on low-rank updates of a physics-based linearization.
    \item After transforming the learned model into an equivalent set of MILP constraints, we use iterative bound tightening, ReLU pruning, and parameter matrix sparsification in order to compress the effective size and complexity of the learned power flow model.
    \item We pose a novel formulation of the AC-UC problem, where the AC power flow constraints are directly replaced by the piecewise linear power flow mapping.
    \item Finally, we compare UC solutions using the NN-based piecewise linear power flow approximation to solutions using DC and linearized power flow approximations. Feasibility of the resulting commitment schedules are determined by solving a multi-time period AC-OPF (MTP AC-OPF) problem.
\end{enumerate}
This paper is structured as follows. In Section~\ref{sec:PF_Mapping}, we develop a NN-based piecewise linear power flow mapping based on low-rank linearization updates. We subsequently cast the learned model as a MILP and perform compression. In Section~\ref{sec:UC}, we pose standard UC formulations, and we show how our piecewise linear power flow mapping can be used to replace the standard power flow constraints. Then, in Section~\ref{sec:TestResults}, we test the performance of the learned power flow mapping by comparing UC and MTP AC-OPF solutions collected from \mbox{14-,} \mbox{57-,} and 118-bus test cases. Finally, conclusions are offered in Section~\ref{sec:Conclusions}.

\section{Learning an Optimally Compact\\ Power Flow Mapping}\label{sec:PF_Mapping}
In this section, we first define a standard power flow model. Next, we develop a NN-based piecewise linear mapping of the power flow equations. This mapping is generated by a training procedure which learns optimal low-rank updates of a physics-based linearization. 
The resulting model is then reformulated as a MILP. Finally, we compress this model via bound tightening, ReLU pruning, and matrix sparsification.

\subsection{Statement of Network Model}
Consider a power network with bus set ${\mathcal N}=\{1,2,...,n\}$, line set ${\mathcal L}=\{1,2,...,m\}$, and signed incidence matrix ${\bm E}\in {\mathbb R}^{m\times n}$. The nodal admittance matrix ${\bm Y}_b\in {\mathbb C}^{n\times n}$ relates complex nodal voltages and power injections via 
\begin{align}
p^{\rm inj}+jq^{\rm inj}=ve^{j\theta}\odot\big({\bm Y}_b ve^{j\theta}\big)^{*},
\end{align}
where $v,\theta,p^{{\rm inj}},q^{{\rm inj}}\in\mathbb{R}^{n}$ and Hadamard product $\odot$ performs component-wise multiplication. Complex line flows, in both directions, are related through line matrices $\bm{Y}_{{\rm ft}},\bm{Y}_{{\rm tf}}\in\mathbb{C}^{m\times n}$:
\begin{align}
p^{{\rm ft}}+jq^{{\rm ft}} & =ve^{j\theta}\odot\big(\bm{Y}_{{\rm ft}}ve^{j\theta}\big)^{*} \label{pft_q_ft_def}\\
p^{{\rm tf}}+jq^{{\rm tf}} & =ve^{j\theta}\odot\big(\bm{Y}_{{\rm tf}}ve^{j\theta}\big)^{*}, \label{ptf_q_tf_def}
\end{align}
where $p^{{\rm ft}},q^{{\rm ft}},p^{{\rm tf}},q^{{\rm tf}}\in\mathbb{R}^{m}$. Apparent powers flows are related by $(s^{{\rm ft}})^{2}=(p^{{\rm ft}})^{2}+(q^{{\rm ft}})^{2}$ and $(s^{{\rm tf}})^{2}=(p^{{\rm tf}})^{2}+(q^{{\rm tf}})^{2}$.
\subsection{A Low-Rank Piecewise Linear Power Flow Mapping Model} \label{subsection: low rank model}
In order to mitigate the computational challenge associated with nonlinear power flow constraints, we use a NN to learn a piecewise linear power flow mapping. For notational convenience, we concatenate all of the active and reactive power injection and bidirectional apparent power flow equations into a function $f:\mathbb{R}^{b=2n}\rightarrow\mathbb{R}^{a=2n+2m}$, such that
\begin{align}\label{eq: f_map}
    f(v,\theta)\rightarrow(p^{\rm inj},q^{\rm inj},s^{\rm ft},s^{\rm tf}).
\end{align}
By defining input vector $x=[v^{T},\theta^{T}]^{T}$ and power flow output vector $y_{\rm pf}=[(p^{\rm inj})^{T},(q^{\rm inj})^{T},(s^{\rm ft})^{T},(s^{\rm tf})^{T}]^{T}$, a linearization of (\ref{eq: f_map}) yields $y_{\rm pf}	\approx f({x}_{0})+\bm{J}({x}_0)\Delta{x}$. This may be rearranged to yield an affine transformation from $x$ to $y_{\rm pf}$:
\begin{align}\label{eq: aff_map}
y_{\rm pf}\approx\bm{J}({x}_{0}){x}+\underbrace{{f}({x}_{0})-\bm{J}({x}_{0}){x}_{0}}_{{r}({x}_{0})},
\end{align}
where ${r}({x}_{0})$ is a residual vector. In order to improve upon the predictive accuracy of the affine mapping in (\ref{eq: aff_map}), an associated piecewise linear mapping from $x$ to $y_{\rm pf}$ may be defined via
\begin{align}\label{eq: y_approx}
y_{\rm pf}\approx\begin{cases}
\bm{J}({x}_{0}){x}+{r}({x}_{0}), & {x}\in\mathcal{R}_{0}\\
\bm{J}({x}_{1}){x}+{r}({x}_{1}), & {x}\in\mathcal{R}_{1}\\
\quad\quad\vdots\\
\bm{J}({x}_{q}){x}+{r}({x}_{q}), & {x}\in\mathcal{R}_{q},
\end{cases}
\end{align}
where ${\mathcal R}_0$, ${\mathcal R}_1$, $\ldots$, ${\mathcal R}_q$ represent the distinct regions of the linearization. Constructing a high-fidelity model of the form (\ref{eq: y_approx}) is generally a challenging task, since selecting optimal points of linearization and separating hyperplanes is a nontrivial task. 

Using a NN to directly model (\ref{eq: y_approx}) can also be challenging, since transforming between Jacobian matrices in different regions requires a full-rank correction, i.e., $\bm{J}({x}_{1})=\bm{J}({x}_{0}) +  {\bm W}$, $\bm{W}\in{\mathbb R}^{a\times b}$, ${\rm rank}\{\bm{W}\}={\rm min}(a,b)$, and thus requires a potentially massive number of nonlinear activation functions. However, this correction can often be approximated by a \textit{low-rank} surrogate, $\bm{W}_{{\rm LR}}$, ${\rm rank}\{\bm{W}_{{\rm LR}}\}=\rho\ll{\rm min}(a,b)$. Low-rank matrices can always by decomposed into the outer product of two matrices, denoted here by ${\bm w}_1\in{\mathbb R}^{b\times \rho}$ and ${\bm w}_2\in {\mathbb R}^{a\times \rho}$, such that $\bm{W}_{{\rm LR}}=\bm{w}_{2}\bm{w}_{1}^{T}$. 

In order to control how such low-rank updates are applied to a Jacobian, ReLU activation functions can be used. For example, consider the rank-1 ($\rho=1$) update case; in this case, $\bm{w}_{1}$, $\bm{w}_{1}$ reduce to vectors ${w}_{1}$, ${w}_{2}$. If we define the hyperplane between two adjacent regions, e.g., ${\mathcal R}_0$ and ${\mathcal R}_1$, as $-b={w}_{1}^{T}{x}$, then a piecewise linear prediction ${y}_{\rm pw}$ can be captured using a single ReLU activation function $\sigma(\cdot)$:
\begin{align}\label{eq: signle_relu}
    {{y}}_{\rm pw}=\bm{J}_0{x}+{r}_0+{w}_{2}\sigma({w}_{1}^{T}{x}+b),
\end{align}
where ${r}_{0}\triangleq{r}({x}_{0})$, $\bm{J}_{0}\triangleq\bm{J}({x}_{0})$, etc. When the ReLU is not activated (${w}_{1}^{T}{x}\!\le\!-b$), the affine transformation of ${\mathcal R}_0$ in (\ref{eq: y_approx}) is exactly recovered by (\ref{eq: signle_relu}). However, when the ReLU is activated (${w}_{1}^{T}{x}>-b$), a rank-1 update is naturally applied to ${\bm J}_0$:
\begin{align}
{{y}}_{\rm pw} & =\left\{ \begin{array}{ll}
\bm{J}_{0}{x}+{r}_{0}, & {x}\in\mathcal{R}_{0}\\
\underbrace{(\bm{J}_{0}+{w}_{2}{w}_{1}^{T})}_{\text{rank-1 update}}{x}+\underbrace{{r}_{0}+{w}_{2}b}_{\text{residual update}}, & {x}\in\mathcal{R}_{1}.
\end{array}\right.
\end{align}
Generally, by choosing $\rho>1$, higher-rank updates can be applied across more piecewise linear regions. To capture these updates, we approximate the full-order piecewise linear model stated in (\ref{eq: y_approx}) via the function
\begin{align}\label{eq: NN-PF-map}
{{y}}_{\rm pw}=\bm{J}^{\star}{x}+{r}^{\star}+\bm{w}_{2}\sigma(\bm{w}_{1}^{T}{x}+{b}),
\end{align}
where $\bm{J}^{\star}$ and ${r}^{\star}$ are the Jacobian and residual terms associated with a specified equilibrium point around which we learn low-rank updates. The model (\ref{eq: NN-PF-map}) may be interpreted as the application of low-rank updates to a full-rank, physics-based Jacobian. $\bm{J}^{\star}$ is the derivative of the physical power flow mapping in (\ref{eq: f_map}); therefore, it represents the concatenation of power injection and apparent power flow Jacobians: $\bm{J}^{\star}=[\bm{J}_{pq}^{T}\;\bm{J}_{s,{\rm ft}}^{T}\;\bm{J}_{s,{\rm tf}}^{T}]^{T}$; these are given in the Appendix.

The NN-based model of (\ref{eq: NN-PF-map}) consists of a physics-based affine feedthrough term ($\bm{J}^{\star}{x}+{r}^{\star}$) and a latent transformation term $\bm{w}_{2}\sigma(\bm{w}_{1}^{T}{x}+{b})$. While this latent term only has nonlinear activation functions applied to a single layer, the model can still provide a quantity of piecewise linearization regions which grows exponentially with the number of ReLUs.

\begin{lemma}
If (\ref{eq: NN-PF-map}) contains $\rho$ activation functions, then it can provide up to $q=2^\rho$ distinct piecewise linearization regions.
\begin{proof}
The activation of each ReLU generates a rank-1 update of $\bm{J}^{\star}$ and thus corresponds to a distinct piecewise linear region ${\mathcal R}_i$. The $\rho$ independently controlled binary activation functions of (\ref{eq: NN-PF-map}) can therefore model $2^\rho$ piecewise linear regions.
\end{proof}
\end{lemma}
NNs which contain multiple layers of ReLU activation functions can also provide piecewise linear mappings. In this work, however, since we are targeting low-rank power flow approximations, we employ activation functions only on a single layer. For a given number of model parameters, shallow NNs cannot always achieve the same level of modeling power as a deeper NN. Results from \cite{Grimstad:2019}, however, show that the MILP reformulation of a shallow NN solves faster than the MILP of a deeper NN (of equivalent complexity).

The model (\ref{eq: NN-PF-map}) is trained by first collecting input and output training data sets $\bm{X}$ and $\bm{Y}$, respectively. Next, an unconstrained optimization algorithm trains the NN by solving
\begin{align}\label{eq: loss}
\min_{b,\bm{w}_{1},\bm{w}_{2}}\left\Vert \bm{Y}-(\bm{J}^{\star}\bm{X}+r^{\star}+\bm{w}_{2}\sigma(\bm{w}_{1}^{T}\bm{X}+b))\right\Vert _{2}^{2}.
\end{align}\label{eq: loss}

\subsection{Exact Neural Network Reformulation as MILP} \label{subsectin: exact_NN}
Once trained, the NN-based model (\ref{eq: NN-PF-map}) can be reformulated as an equivalent set of MILP constraints~\cite{Venzke:2021}. Defining intermediate variable $\hat{{z}}=\bm{w}_{1}^{T}{x}+{b}$, the ReLU function ${z}=\sigma(\hat{{z}})\triangleq{\rm max}(\hat{{z}},0)$ is captured by the constraints
\begin{equation}\label{eq: ReLU_constraints}
\begin{aligned}
{z}_{i} & \leqslant\hat{{z}}_{i}-M^{{\rm min}}_{i}(1-{\beta}_{i}), & \quad {z}_{i} & \geqslant\hat{{z}}_{i}\\
{z}_{i} & \leqslant M^{{\rm max}}_{i}{\beta}_{i}, & \quad {z}_{i} & \geqslant0,
\end{aligned}
\end{equation}
where $M^{\rm min}_i$ and $M^{\rm max}_i$ are the minimum and maximum values that $\hat{z}^i_k$ can take, respectively, and ${\beta}$ is a vector of binaries: ${\beta}_i \in \{0,1\}$. The tightness of these big-M bounds influence the efficiency of the branch-and-bound algorithm used to handle these constraints~\cite{Grimstad:2019}. With this formulation, the power flow mapping of (\ref{eq: NN-PF-map}) can be exactly captured via
\begin{subequations} \label{eq: NN_MILP}
\begin{align}
{{y}}_{\rm pw}& = \bm{J}^{\star}{x}+{r}^{\star}+\bm{w}_{2}{z}\label{eq: yb}\\
\hat{{z}}& = \bm{w}_{1}^{T}{x}+{b}\\
 \eqref{eq: ReLU_constraints}&, \;\forall i \in \{1,\ldots,\rho\}.\label{eq: 8ae}
\end{align}
\end{subequations}
Neglecting the NN, (\ref{eq: NN_MILP}) reduces to a \textit{linear} power flow model:
\begin{align} \label{eq: lin_model}
{y}_{\rm lin} = \bm{J}^{\star}{x}+{r}^{\star}.
\end{align}

\subsection{Compression of the NN-Based Power Flow Mapping}
Once (\ref{eq: NN-PF-map}) has been reformulated as a MILP, it can be embedded as a constraint into a variety of optimization problems. To limit the computational complexity of the associated mixed integer constraints, however, we iteratively (i) sparsify the NN weighting matrices, (ii) tighten the big-M constraints associated with reformulation (\ref{eq: ReLU_constraints}), and (iii) prune the NN's ReLUs. Each of these steps is summarized below.

\subsubsection{NN Sparsification}
Following the general procedure outlined in \cite{Venzke:2021}, we sparsify the NN weighting matrices ${\bm w}_1$ and ${\bm w}_2$ by setting some targeted percentage of the weights to 0. The weights selected are the ones which have the smallest absolute magnitude. After sparsification, the network is retrained; during retraining, sparsified entries are fixed to $0$.

\subsubsection{Big-M Bound Tightening}
To tighten the big-M bounds, we first define inequality constraints associated with the NN inputs. That is, we define relevant nodal voltage and phase angle inequality constraints (e.g., ${V}_{{\rm min}}\le{ v}_{i}\le{V}_{{\rm max}}$ and $|\theta_{i}-\theta_{j}|\le\Delta\theta_{{\rm max}}$), denoted by ${\mathcal C}_{v}$ and ${\mathcal C}_{\theta}$. Next, we define inequality constraints associated with NN outputs, i.e., power injections and line flow limits, denoted by ${\mathcal C}_{p}$, ${\mathcal C}_{q}$, and ${\mathcal C}_{s}$. The lower bound $M^{\rm min}_i$ can be directly computed via the MILP
\begin{subequations}\label{eq: M_bound_MILP}
\begin{align}
\!\!M^{{\rm min}}_{i}\!=\!\!\min_{{x},{{y}_{\rm pw}},{\beta}}\;\; & \hat{{z}}_{i}\label{eq: M_min}\\
{\rm s.t.}\;\; & {\eqref{eq: yb}-\eqref{eq: 8ae}}\\
 {x}\in & \{{ v},{\theta}\,|\,{v},\theta \in \mathcal{C}_{{ v}},\mathcal{C}_{\theta}\}\\
 {{y}_{\rm pw}}\in & \{{p}^{\rm inj},{q}^{\rm inj},{s}_{l}\,|\,{p}^{\rm inj},{q}^{\rm inj},{s}_{l} \in \mathcal{C}_{p},\mathcal{C}_{q},\mathcal{C}_{s}\}
\end{align}
\end{subequations}
where ${s}_{l}$ denotes the apparent power line flows in both directions. The upper bound $M^{\rm max}_i$ may be computed by maximizing (\ref{eq: M_min}), rather than minimizing it. We note that the constraint sets $\mathcal{C}_{{v}},\mathcal{C}_{\theta},\mathcal{C}_{p},\mathcal{C}_{q},\mathcal{C}_{s}$ can be naively defined using the engineering constraints associated with whatever optimization problem the NN is ultimately being used to solve (e.g., UC, OPF, etc.). Alternatively, these constraint sets can themselves be first tightened using, e.g., optimization-based bound tightening~\cite{Coffrin:2018} or analytic methods~\cite{Shchetinin:2019}.

\subsubsection{ReLU Pruning}
Using the calculated big-M bounds, individual ReLUs can be pruned (i.e., removed) from the NN. Pruning procedure: if $M^{{\rm max}}_{i}\le0$, then the associated ReLU is never active, and ${\beta}_i$ is fixed to $0$. However, if $\;M^{{\rm min}}_{i}>0$, then the ReLU is always active, and ${\beta}_i$ is fixed to $1$.

\section{Unit Commitment Formulations}\label{sec:UC}

In this section, we first present the three-binary formulation of the AC-UC problem, which is introduced in \cite{morales2012tight}, using the indexing and notation schemes in \cite{liu2018global}.
We then present the UC problem where we approximate the AC power flows using the novel NN-based piecewise linear model, which has been exactly reformulated as the set of MILP constraints given in Section~\ref{subsectin: exact_NN}.
Next, we present two popular power flow approximations for comparison and benchmarking purposes. 
First, the AC-UC problem with linearized power flow equations and second, the DC-UC problem, where we neglect reactive power, line losses, and voltage magnitude deviations.
Last, we present the MTP AC-OPF formulation.
The UC schedules found using the three power flow approximation methods are tested for AC feasibility through checking for the feasibility of their corresponding MTP AC-OPF solutions.

\subsection{Objective and cost constraints}

Let $\mathcal{T} = \{1, \ldots, T\}$ be the set of time indices, where each time index represents an hour of simulation and $T$ is the total simulation time.
Let $\mathcal{G} = \{1, \ldots, G\}$ be the set of generators.
For each generator $g \in \mathcal{G}$ and each time period $t \in \mathcal{T}$, there is a binary variable $y_{g,t} \in \{0, 1\}$ indicating whether the generator is ON ($y_{g,t} = 1$) or OFF ($y_{g,t} = 0$).
Let $y = \{y_{g,t} \ | \ g \in \mathcal{G}, t \in \mathcal{T} \}$ be the set of all generator statuses. 
Let $u_{g,t} \in \{0,1\}$ and $w_{g,t}\in \{0,1\}$ be the start-up and shut-down statuses, respectively, of generator $g \in \mathcal{G}$ at time $t \in \mathcal{T}$. 
If $u_{g,t}=1$, then generator $g$ starts-up at the beginning of hour~$t$ and is zero otherwise. 
If $w_{g,t}=1$, then generator $g$ shuts-down at the beginning of hour $t$ and is zero otherwise.
Let $u = \{u_{g,t} \ | \ g \in \mathcal{G}, t \in \mathcal{T} \}$ and $w = \{w_{g,t} \ | \ g \in \mathcal{G}, t \in \mathcal{T} \}$.

Let $c(\cdot)$ be the total cost of operating the network:
\begin{align}
    c(p^\Delta, u,w) = c^p(p^\Delta) + c^{su}(u,w), \label{cost}
\end{align}
where $c^p(\cdot)$ is the production cost and $c^{su}(\cdot)$ is the start-up cost.
Let $p^\Delta_{g,t} \geqslant 0$ be the real power production of generator $g \in \mathcal{G}$ at time $t \in \mathcal{T}$ above $P_g^{\text{min}} \geqslant 0$, the minimum real power production limit for generator $g$.
Let $p^\Delta = \{p^\Delta_{g,t} \ | \ g \in \mathcal{G}, t \in \mathcal{T} \}$.
The production cost $c^p(\cdot)$ is:
\begin{align}
    c^p(p^\Delta) = \sum\nolimits_{g \in \mathcal{G}} \sum\nolimits_{t \in \mathcal{T}} c^p_{g,t}(p^\Delta_{g,t}),
\end{align}
where $c^p_{g,t}(\cdot)$ is a convex piecewise linear function for all $g \in \mathcal{G}$ and $t \in \mathcal{T}$.
The start-up cost $c^{su}(\cdot)$ has the form: 
\begin{align}
    c^{su}(u,w) = \sum\nolimits_{g \in \mathcal{G}} \sum\nolimits_{t \in \mathcal{T}} c^{su}_{g,t}(u,w),
\end{align}
where $c^{su}_{g,t}(\cdot)$ is a monotonically increasing step function representing the startup costs for generator $g \in \mathcal{G}$, which increase with the amount of time the generator has been shut-down.
See \cite{liu2018global} for more details on this formulation.

\subsection{Generation Constraints}

For generator $g\in \mathcal{G}$, let the minimum uptime (the minimum period of time that the generator must be online before changing status) be $T_g^u$, and the minimum downtime (the minimum period of time that the generator must be offline before changing status) is $T_g^d$.
These constraints are given by: 
\begin{align}
    &\sum\nolimits^t_{t' = t - T_g^u + 1} u_{g,t'} \leqslant y_{g,t} & \forall g \in \mathcal{G}, t \in \mathcal{T} \label{minimum_uptime}\\
    &\sum\nolimits^t_{t' = t - T_g^d + 1} w_{g,t'} \leqslant 1-y_{g,t} & \forall g \in \mathcal{G}, t \in \mathcal{T}. \label{minimum_downtime}
\end{align}
For values of $t' < 1$ (i.e., before the start of the simulation), we assume the values of $u_{g,t'}$ and $w_{g,t'}$ are known parameters.

Generators cannot start-up and shut-down in the same time period.
This is enforced via \eqref{minimum_uptime} and \eqref{minimum_downtime} in conjunction with:
\begin{align}
    y_{g,t} - y_{g,t-1} = u_{g,t} - w_{g,t}  \qquad \forall g \in \mathcal{G}, t \in \mathcal{T}. \label{startup_shutdown}
\end{align}

Let $r_{g,t} \geqslant 0$ be the real power reserve available to generator $g \in \mathcal{G}$ at time $t \in \mathcal{T}$, and let $ r = \{r_{g,t} \ | \ g \in \mathcal{G}, t \in \mathcal{T} \}$.
Parameter $P^R_t$ is the total spinning reserve needed for the network at $t \in \mathcal{T}$, and we require that:
\begin{align}
    P^R_t \leqslant \sum\nolimits_{g \in \mathcal{G}} r_{g,t}  \qquad t \in \mathcal{T}. \label{spinning_reserve}
\end{align}

Let parameter $P_g^{\text{max}} \geqslant 0$, be the maximum real power production limit for generator $g \in \mathcal{G}$. 
Parameters $SU_g$ and $SD_g$ are the maximum real power a generator $g \in \mathcal{G}$ can produce immediately after starting up, and immediately before shutting down, respectively.
Then, assuming $SU_g, SD_g \leqslant P_g^{\text{max}}$, the start-up generation limits are enforced via the following constraints:
\begin{align}
        p^{\Delta}_{g,t} + r_{g,t} &\leqslant (P_g^{\text{max}} - P_g^{\text{min}})y_{g,t} - (P_g^{\text{max}} - SU_g)u_{g,t}\nonumber\\
     - (P_g^{\text{max}} -& SD_g)w_{g,t+1},  \forall g \in \{ i \in \mathcal{G} \ | \ T_i^u \geqslant 2  \}, t \in \mathcal{T} \label{start_up}\\
     p^{\Delta}_{g,t} + r_{g,t} &\leqslant (P_g^{\text{max}} - P_g^{\text{min}})y_{g,t} - (P_g^{\text{max}} - SU_g)u_{g,t}\nonumber\\
     &\;\;\quad\qquad\qquad \forall g \in \{ i \in \mathcal{G} \ | \ T_i^u = 1  \}, t \in \mathcal{T} \label{start_up}\\
     p^{\Delta}_{g,t} &\leqslant (P_g^{\text{max}} - P_g^{\text{min}})y_{g,t} - (P_g^{\text{max}} - SD_g)w_{g,t+1}\nonumber\\
     &\;\;\quad\qquad\qquad \forall g \in \{ i \in \mathcal{G} \ | \ T_i^u = 1  \}, t \in \mathcal{T}. \label{shut_down}
\end{align}
We also require a constraint to ensure the shut-down generation limits are enforced during the first time period $t=1$:
\begin{align}
    w_{g,1} \leqslant 0  \qquad \forall g \in \{ \ i \in \mathcal{G} \ | \ P_{i}^{\text{init}} > SD_i\}, \label{shut_down_init}
\end{align}
where $P_{i}^{\text{init}}$ is the real power produced by generator $g \in \mathcal{G}$ the time period before the simulation begins. Parameters $RU_g$ and $RD_g$ are the real power ramp-up and ramp-down limits, respectively, for generator $g \in \mathcal{G}$.
Generators must abide by ramping limits, which restrict the change in real power:
\begin{align}
    p^\Delta_{g,t} + r_{g,t} - p^\Delta_{g,t-1} &\leqslant RU_g \qquad \forall g \in \mathcal{G}, t \in \mathcal{T} \label{ramp_up}\\
    -p^{\Delta}_{g,t} + p^{\Delta}_{g,t-1} &\leqslant RD_g \qquad \forall g \in \mathcal{G}, t \in \mathcal{T}. \label{ramp_down}
\end{align}

Generator $g \in \mathcal{G}$ has lower and upper reactive power limits, $Q_g^{\text{min}}$ and $Q_g^{\text{max}}$, respectively.
Let $q_{g,t}$ be the reactive power production of generator $g \in \mathcal{G}$ at time $t \in \mathcal{T}$.
The reactive power output of each generator is constrained to be within its limits when active: 
\begin{align}
    Q_g^{\text{min}}y_{g,t} \leqslant q_{g,t} \leqslant Q_g^{\text{max}}y_{g,t} \qquad \forall g \in \mathcal{G}, t \in \mathcal{T}. \label{reactive_generation_limits}
\end{align}

\subsection{AC-OPF Constraints}
Recall that ${\mathcal N}$ is the set of buses in the network.
Now, let $\mathcal{G}_b$ be the set of generators at bus $b \in \mathcal{N}$. 
Parameters $P^D_{b,t}$ and $Q^D_{b,t}$ are the real and reactive power demand at $b \in \mathcal{N}$ and $t \in \mathcal{T}$.
Let $p^{\text{inj}}_{b,t}$ and $q^{\text{inj}}_{b,t}$ be the real and reactive power, respectively, \textit{injected into the network} from bus $b \in \mathcal{N}$ at time $t \in \mathcal{T}$.
Then, the real power balance constraints are:
\begin{align}
    p^{\text{inj}}_{b,t} = \sum_{g \in \mathcal{G}_b} \left(p^\Delta_{g,t} + P_g^{\text{min}}y_{g,t}\right)-P^D_{b,t}  \quad \forall b \in \mathcal{N}, t \in \mathcal{T}, \label{real_power_balance}
\end{align}
and the reactive power balance constraints are:
\begin{align}
    q^{\text{inj}}_{b,t} = \sum_{g \in \mathcal{G}_b} q^G_{g,t} + \sum_{g \in \mathcal{SC}_b} q^{SC}_{g,t}-Q^D_{b,t}  \quad \forall b \in \mathcal{N}, t \in \mathcal{T}. \label{reactive_power_balance}
\end{align}

Recall that $\mathcal{L}$ is the set of transmission lines in the network.
Each line $\ell \in \mathcal{L}$ has a designated ``from'' bus and ``to'' bus, which can be arbitrarily chosen.
$s^{\text{ft}}_{\ell,t}$ is the apparent power flow on line $\ell \in \mathcal{L}$ at time $t \in \mathcal{T}$ with flow from the ``from'' bus and to the ``to'' bus; $s^{\text{tf}}_{\ell,t}$ is defined oppositely.
The apparent power flows on each line $\ell \in \mathcal{L}$ cannot exceed their maximum allowable limit $S_{l}^{\text{max}}$:
\begin{align}
    s^{\text{ft}}_{\ell,t} \leqslant S_{l}^{\text{max}}, \quad  s^{\text{tf}}_{\ell,t} \leqslant S_{l}^{\text{max}} \qquad \forall l \in \mathcal{L}, t \in \mathcal{T}. \label{apparent_power}
\end{align}

Let $v_{b,t}$ be the voltage magnitude at bus $b \in \mathcal{N}$ and time $t \in \mathcal{T}$.
Parameters $V_b^{\text{max}}$ and $V_b^{\text{min}}$ are the maximum and minimum voltage magnitude limits, respectively, for bus $b \in \mathcal{N}$.
Then, the constraints on the voltage magnitudes are as follows:
\begin{align}
    V_b^{\text{min}} \leqslant v_{b,t} \leqslant V_b^{\text{max}} \qquad \forall b \in \mathcal{N}, t \in \mathcal{T}. \label{voltage_magnitude_limits}
\end{align}
We designate the ``from'' and ``to'' bus voltage angles of line $\ell \in \mathcal{L}$ as $\theta_{\ell_{\rm f}}$ and $\theta_{\ell_{\rm t}}$, respectively. Parameters $\Theta^{min}_\ell$ and $\Theta^{max}_\ell$ are the minimum and maximum voltage angle differences for line $\ell \in \mathcal{L}$.
We constrain the voltage angle differences as:
\begin{align}
    \Theta^{min}_\ell \leqslant \theta_{\ell_{\rm f}}-\theta_{\ell_{\rm t}} \leqslant \Theta^{max}_\ell \qquad \forall \ell \in \mathcal{L}, t \in \mathcal{T}. \label{voltage_angle_diff_limits}
\end{align}
Last, we set the voltage angle of the reference bus to zero for all $t \in \mathcal{T}$.
Let $b_{\text{ref}} \in \mathcal{N}$ designate the reference bus. Then:
\begin{align}
    \theta_{b_{\text{ref}},t} = 0 \qquad \forall t \in \mathcal{T}. \label{voltage_angle_ref}
\end{align}

\subsection{AC-UC Formulation}
Let $p^{\text{ft}}_{\ell,t}$, $p^{\text{tf}}_{\ell,t}$, $q^{\text{ft}}_{\ell,t}$, and $q^{\text{tf}}_{\ell,t}$ be the the real and reactive power flows on lines $\ell \in \mathcal{L}$ at time $t \in \mathcal{T}$, which are defined in vector form in \eqref{pft_q_ft_def} and \eqref{ptf_q_tf_def}.
Now, let $\mathcal{L}^{\rm ft}_b \subseteq \mathcal{L}$ and $\mathcal{L}^{\rm tf}_b \subseteq \mathcal{L}$ be the subsets of lines that are originating and terminating, respectively, at bus $b \in \mathcal{N}$.
Parameters $G_b^{\text{sh}}$ and $B_b^{\text{sh}}$ are the shunt conductance and susceptance, respectively, at bus $b \in \mathcal{N}$.
Then, the power injected into the network from bus $b \in \mathcal{N}$ is equivalent to:
\begin{align}p_{b,t}^{\text{inj}} & =G_{b}^{\text{sh}}v_{b,t}^{2}+\sum\nolimits_{\ell\in\mathcal{L}_{b}^{{\rm ft}}}{p_{\ell,t}^{\text{ft}}}+\sum\nolimits_{\ell\in\mathcal{L}_{b}^{{\rm tf}}}{p_{\ell,t}^{\text{tf}}}\\
q_{b,t}^{\text{inj}} & =-B_{b}^{\text{sh}}v_{b,t}^{2}+\sum\nolimits_{\ell\in\mathcal{L}_{b}^{{\rm ft}}}q_{\ell,t}^{\text{ft}}+\sum\nolimits_{\ell\in\mathcal{L}_{b}^{{\rm tf}}}q_{\ell,t}^{\text{tf}}.
\end{align}
The apparent power flows are equal to:
\begin{align}
     s^{\text{ft}}_{\ell,t} &= \sqrt{(p^{\text{ft}}_{\ell,t})^2 + (q^{\text{ft}}_{\ell,t})^2} \qquad \forall \ell \in \mathcal{L}, t \in \mathcal{T}\label{S_fr_pq}\\
     s^{\text{tf}}_{\ell,t} &= \sqrt{(p^{\text{tf}}_{\ell,t})^2 + (q^{\text{tf}}_{\ell,t})^2} \qquad \forall \ell \in \mathcal{L}, t \in \mathcal{T}. \label{S_to_pq}
\end{align}
Then, the MINLP formulation of the AC-UC problem is:
\begin{equation}
\begin{aligned}
\min\limits_{\mathcal{X}^\text{cont}, \mathcal{X}^\text{bin},\theta, v, p^{\text{ft}}, p^{\text{tf}}, q^{\text{ft}}, q^{\text{tf}}}& \  \eqref{cost} \ \ \text{s.t.} \ \eqref{minimum_uptime} - \eqref{S_to_pq}, \nonumber
\end{aligned}
\tag{AC-UC}\label{AC-UC}
\end{equation}
where we collect all the continuous generation variables in the set $\mathcal{X}^\text{cont} = \{ r_{g,t}, p^{\Delta}_{g,t}, q_{g,t}, q^{SC}_{g,t} \ | \ g \in \mathcal{G}, t \in \mathcal{T} \}$, and all the binary commitment decision variables in the set $\mathcal{X}^\text{bin} = \{ y_{g,t}, u_{g,t}, w_{g,t} \ | \ g \in \mathcal{G}, t \in \mathcal{T} \}$.
Note that each variable corresponding to the individual elements of vectors $p^{\text{ft}}$, $p^{\text{tf}}$, $q^{\text{ft}}$, $q^{\text{tf}}$, $v$, and $\theta$ are optimization variables as well.

\subsection{UC using power flow approximations}

We model power flow using the piecewise linear NN model described in Section~\ref{subsection: low rank model}. The corresponding MILP model is presented in \eqref{eq: NN_MILP}.
Then, the NN-based AC-UC problem is:
\begin{equation}
\begin{aligned}
\min\limits_{\mathcal{X}^\text{cont}, \mathcal{X}^\text{bin}, \theta, v}& \quad  \eqref{cost} \ \ \text{s.t.} \ \eqref{eq: NN_MILP},\, \eqref{minimum_uptime} - \eqref{voltage_angle_diff_limits}. \nonumber
\end{aligned}
\tag{NN AC-UC}\label{NN-AC-UC}
\end{equation}


A linearized power flow model is presented in \eqref{eq: lin_model}.
Then, the following MILP is the version of the AC-UC problem using this linearized power flow model:
\begin{equation}
\begin{aligned}
\min\limits_{\mathcal{X}^\text{cont}, \mathcal{X}^\text{bin}, v, \theta}& \quad  \eqref{cost} \ \ \text{s.t.} \ \eqref{eq: lin_model}, \, \eqref{minimum_uptime} - \eqref{voltage_angle_diff_limits}. \nonumber
\end{aligned}
\tag{L AC-UC}\label{linear-AC-UC}
\end{equation}

In the DC-UC problem, we neglect reactive power and line losses, and voltage magnitudes deviations. 
It then follows that the real and apparent power flows are:
\begin{align}
    p^{\text{ft}}_{\ell, t} &= -p^{\text{tf}}_{\ell, t} \qquad\qquad\qquad \forall \ell \in \mathcal{L}, t \in \mathcal{T} \label{pft_equal_ptf}\\
    p^{\text{ft}}_{\ell, t} &= -B_\ell(\theta_{\ell_{\rm f}}-\theta_{\ell_{\rm t}})  \qquad \forall \ell \in \mathcal{L}, t \in \mathcal{T} \label{p_DC}\\
     s^{\text{ft}}_{\ell,t} &= p^{\text{ft}}_{\ell, t},  \ s^{\text{tf}}_{\ell,t} = -p^{\text{ft}}_{\ell, t} \quad\,\forall \ell \in \mathcal{L}, t \in \mathcal{T}. \label{S_DC}
\end{align}
Now, we define the DC-UC problem as:
\begin{equation}
\begin{aligned}
&\min\limits_{\mathcal{X}^\text{cont}, \mathcal{X}^\text{bin}, \theta, p^{\text{ft}}, p^{\text{tf}}} \quad  \eqref{cost}\\
& \ \text{s.t.} \ \eqref{minimum_uptime} - \eqref{ramp_down}, \eqref{real_power_balance}, \eqref{apparent_power}, \eqref{voltage_angle_diff_limits}, \eqref{voltage_angle_ref},  \eqref{pft_equal_ptf} - \eqref{S_DC}. \nonumber
\end{aligned}
\tag{DC-UC}\label{DC-UC}
\end{equation}

\subsection{Multi-time period AC-OPF formulation} \label{sec:MTP}

When formulating the MTP AC-OPF problem, we assume  all elements in $\mathcal{X}^\text{bin}$ are known parameters, i.e., the commitment schedule is set, and we optimize over continuous generation variables $\mathcal{X}^\text{cont}$.
The MTP AC-OPF problem follows:
\begin{equation}
\begin{aligned}
\min\limits_{\mathcal{X}^\text{cont}, \theta, v, p^{\text{ft}}, p^{\text{tf}}, q^{\text{ft}}, q^{\text{tf}}}& \  \eqref{cost} \ \ \text{s.t.} \ \eqref{minimum_uptime} - \eqref{S_to_pq}. \nonumber
\end{aligned}
\tag{MTP AC-OPF}\label{MTP AC-OPF}
\end{equation}

The MTP AC-OPF problem is used to test the feasibility of the commitment schedules resulting from solving the NN AC-UC, L AC-UC, and DC-UC problems.
We classify a UC solution that results in an infeasible MTP AC-OPF problem as a \textit{infeasible commitment schedule}, i.e., the selection of binary variables in set $\mathcal{X}^{\text{bin}}$ cannot be realized.

\section{Test Results}\label{sec:TestResults}
In this section, we present results collected on the \mbox{14-}, \mbox{57-}, and 89-bus PGLib-OPF test cases~\cite{Babaeinejadsarookolaee:2019} over a 24-hour period. In Section~\ref{sec: compact_vs_direct}, we first compare the expressive power of the compact NN model (\ref{eq: NN-PF-map}) to a direct power flow mapping, which is exclusively used in the literature. In Section~\ref{sec: UC_tests}, we use the compact NN models to solve the NN AC-UC problems; then, we compare the obtained solutions to those generated by the linear benchmarks.

\subsection{Expressive Power of the Compact NN}\label{sec: compact_vs_direct}
\begin{figure}
    \centering
    \includegraphics[width=1\columnwidth]{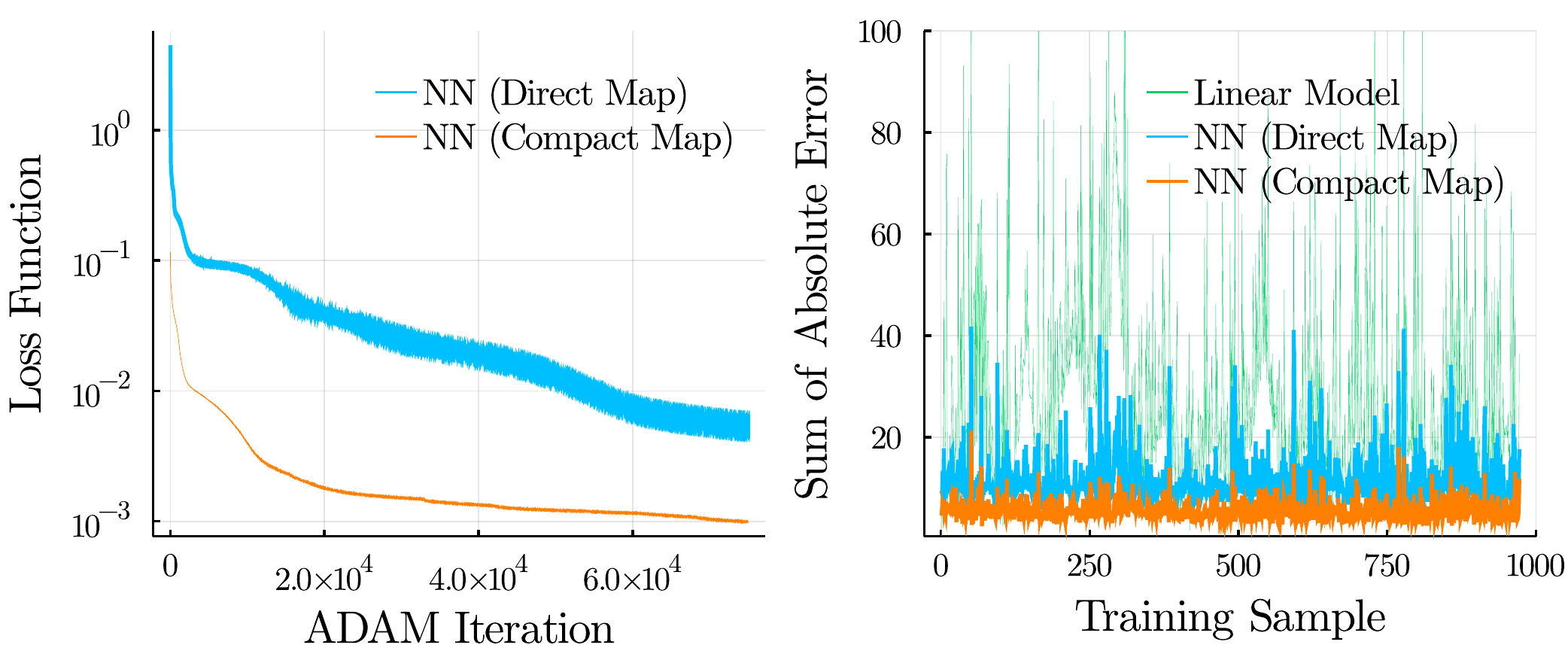}
    \caption{The left panel shows the loss function of the compact the direct NN mappings during training. The right panel shows linear, direct NN, and compact NN per-unitized error ($\left\Vert y_{\rm pf}-y_{{\rm lin}}\right\Vert_{1}$, $\left\Vert y_{\rm pf}-y_{{\rm nn}}\right\Vert_{1}$, and $\left\Vert y_{\rm pf}-y_{{\rm pw}}\right\Vert_{1}$, respectively)  associated with the the 89-bus system.}
    \label{fig:Compare_NNs}
\end{figure}

In order to test the expressive power of the compact NN, we collected 972 feasible power flow solutions from the 89-bus system; loads were chosen by looping over Unit Commitment load curves (see the following subsection for more details regrading data collection). We then trained a compact NN power flow mapping of the form (\ref{eq: NN-PF-map}) with $\rho = 25$ ReLUs using ADAM in Flux. Data were shuffled and mini-batched into sets of 75, and a learning rate of $\eta=2.5\times10^{-4}$ was used. A second NN was also trained, but this model mapped power flow inputs ($v$ and $\theta$) \textit{directly} to power flow outputs via ${{y}}_{\rm nn}=\bm{w}_{2}\sigma(\bm{w}_{1}^{T}{x}+{b})$, which we refer to as a ``direct" NN mapping; this model was also trained with 25 ReLUs. Both NNs were allowed to train for $7.5\times10^4$ steps, where loss functions minimization showed signs of saturation. Results are shown in Fig. \ref{fig:Compare_NNs}, where the left panel shows loss function saturation, and the right panel depicts the predictive accuracy of the models. For reference, we also plot the prediction of linear power flow model ${y}_{\rm lin}=\bm{J}^{\star}{x}+{r}^{\star}$ from (\ref{eq: lin_model}). NN power flow predictions are generally an order of magnitude better than linear model predictions, and the compact NN predictions are generally over a factor of two better than the direct NN.

\subsection{Data Collection and NN Training}\label{sec: UC_tests}
We used the UC nodal load curves and generation cost curves developed in the UnitCommitment.jl package~\cite{Xavier:2021} for these systems. Reactive power load curves were generated for each system by assuming constant power factors at each load. For increased complexity, we assumed the active power limits for each generator in the 14 and 57 bus systems matched those of the associated M{\sc atpower} test cases (i.e., contrary to PGLib-OPF, M{\sc atpower} assumes that no generating unit acts only as a synchronous condenser; instead, every generator can produce active power and, therefore, a larger number of generators can participate in UC). We also decreased the apparent power thermal limits in these systems by 30\%. 

The power flow mapping \eqref{eq: NN-PF-map} can potentially be trained on any set of feasible power flow solutions. To collect training data in a targeted way, we first gathered the hourly load profiles associated with the UC problems. For each hour, we used PowerModels.jl~\cite{Coffrin:2018} to generate a \textit{feasible} power flow solution (i.e., a power flow solution which satisfied all thermal, voltage, and generation limit constraints). For each hour, we also generated feasible power flow solutions with each generator turned off, and with random combinations of up to three other generators also turned off; infeasible samples were rejected. In order to sample over a larger feasible space, we also perturbed the generator voltage limits at each power flow solve. That is, we randomly ``pushed" the ${V}^{\rm min}$ and ${V}^{\rm max}$ generator constraints up and down, respectively, to nonstandard values which were still within the feasible space.

Once training and testing data sets were collected, compact NN-based power flow mapping models were trained and compressed. NN models of the \mbox{14-}, \mbox{57-}, and 89-bus power systems were trained on 219, 532, and 972 power flow samples, respectively. Notably, each training sample  contained $2n-1$ inputs (reference bus phase angle was never included as an input) and $2n+2m$ outputs, according to mapping (\ref{eq: f_map}). 
The NNs associated with these systems contained 20, 25, and 30 ReLUS, respectively. Models were trained using ADAM in Flux, learning rates were set between $(1-3)\times 10^{-4}$, and all data were shuffled and mini-batched.

\subsection{Unit Commitment Experiment Results}\label{sec: results}

\begin{figure}
    \centering
    \includegraphics[width=1\columnwidth]{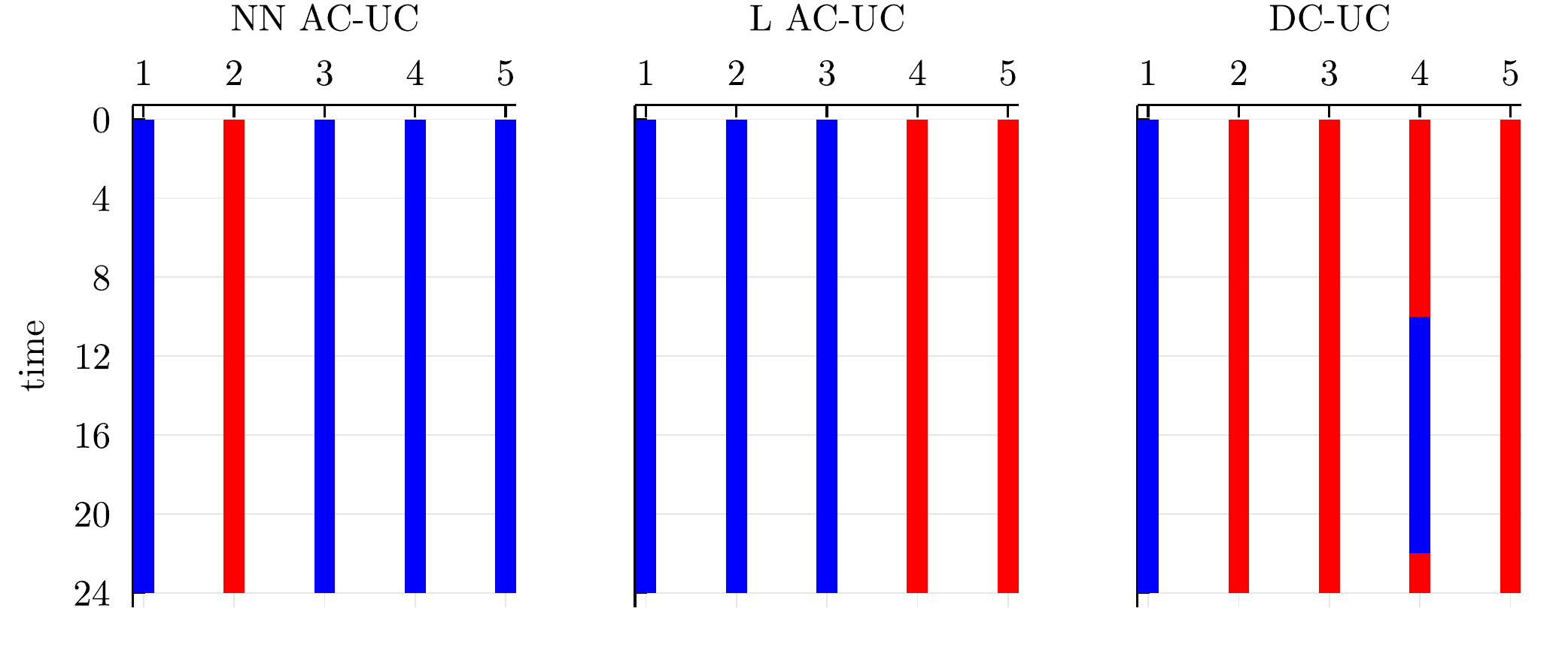}
    \caption{These three plots show the unit status decisions for the 14-bus network resulting  from solving the NN AC-UC, L AC-UC and DC-UC problems. The top labels ($1-5$) refer to the unit numbers, and the left axis marks a 24-hour period. Blue sections represent time periods where the status of the corresponding unit is ON, while red sections represent OFF periods.}
    \label{fig:14_bus_unit_decisions}
    \vspace{-1mm}
\end{figure}

We solved the three versions of the UC problem discussed in Section \ref{sec:UC} (NN AC-UC, L AC-UC, and DC-UC) for the three considered networks (14-, 57-, and 89-bus). For each system, the linearization terms ($\bm{J}^{\star}$, ${r}^{\star}$) were generated from a power flow solution of the UC hour-1 (mean value) base load level with all generators turned ON and producing. Figure \ref{fig:14_bus_unit_decisions} shows the resulting unit commitment decisions for the 14-bus network.
Blue and red lines mark time periods when the corresponding unit is ON and OFF, respectively.
Here, we clearly see different unit commitment schedules are chosen.
Note that the NN-based solution turns on more generators than the other power flow approximations.
Across the three tested networks, the commitment schedules resulting from the NN-based and linear formulations are MTP AC-OPF feasible (see Section \ref{sec:MTP}) in the base-loading case.  
However, the committment schedules found via DC-UC only resulted in a feasible MTP AC-OPF for the 57-bus network.

Next, we tested the performance of the UC formulations using three different load alteration schemes with the goal of capturing various multi-time loading possibilities. 
In the first scheme, we uniformly scaled all real and reactive loads in the network over all time periods using the same scaling value.
In the second scheme, we scaled the loads at each bus over all time periods using a randomly sampled value.
The last load alteration scheme aimed to increase the peak loads and decrease the lowest loads, thus forcing more time-varying commitment decisions.
We accomplished this by scaling each load at each hour $t$ according to: $1 + a \sin(\frac{2\pi}{24t})$, where $a$ is the amplitude of the sine.
We tested 10 cases for each loading scheme, for a total of 30 varied loading scenarios, and constrained our scaling range to $\pm 15\%$ of the original loads.
For the first and last loading schemes, we tested samples evenly distributed within this scaling range.

Table \ref{results_table} summarizes the MTP AC-OPF feasibility results, which use the binary variables from the UC solutions shown in the left-most column. 
Each UC formulation has a tallied number of total feasible and infeasible solutions, as well as the number of scenarios that did not produce a solution (either due to an infeasible UC problem or the inability of the MTP AC-OPF problem to converge within a reasonable time limit).
The UC problems were run until a relative MIP gap of 1\% was achieved; otherwise, the best feasible solution found after one hour of solving was used.

Overall, the NN-based method outperforms the linear and DC approximations.
There are multiple cases where the NN-based approximation is the \textit{only} formulation that selects a feasible unit commitment schedule.
For the 14-bus network, this is true for 15 out the the total 30 test cases, and for the 89-bus network, 5 out of the 30 test cases.
Furthermore, the NN-based commitment schedules are \textit{only} MTP AC-OPF infeasible (or unable to find a solution) when both the linear and DC methods are infeasible as well.
Also, note that the feasibility of the test cases were not verified.

For the 57-bus network, we found that the linear power-flow approximation performed equally as well as the NN-based approximation in all cases.
Most unit commitment schedules for this test case required all generators to be ON at all times, which, we hypothesize, did not require the additional accuracy afforded by the NN-based UC formation.

\begin{table}[]
\caption{MTP AC-OPF results for load variations}
\begin{center}
\begin{tabular}{cl|c|c|c|}
\cline{3-5}
\multicolumn{1}{l}{}                      &            & 14-bus & 57-bus & 89-bus \\ \hline
\multicolumn{1}{|c|}{\multirow{3}{*}{NN AC-UC}} & Feasible  & 28 & 30 & 28 \\ 
\multicolumn{1}{|c|}{}                    & Infeasible & 2  & 0  & 1  \\ 
\multicolumn{1}{|c|}{}                    & No solution        & 0  & 0  & 1  \\ \hline
\multicolumn{1}{|c|}{\multirow{3}{*}{L AC-UC}}  & Feasible   & 13 & 30 & 22 \\ 
\multicolumn{1}{|c|}{}                    & Infeasible & 15 & 0  & 8  \\ 
\multicolumn{1}{|c|}{}                    & No solution         & 2  & 0  & 0  \\ \hline
\multicolumn{1}{|c|}{\multirow{3}{*}{DC-UC}} & Feasible   & 0  & 23 & 19 \\ 
\multicolumn{1}{|c|}{}                    & Infeasible & 30 & 7  & 10  \\ 
\multicolumn{1}{|c|}{}                    & No solution       & 0  & 0  & 1  \\ \hline
\end{tabular}
\label{results_table}
\end{center}
\vspace{-6mm}
\end{table}

\begin{figure}
    \centering
    \includegraphics[width=1\columnwidth]{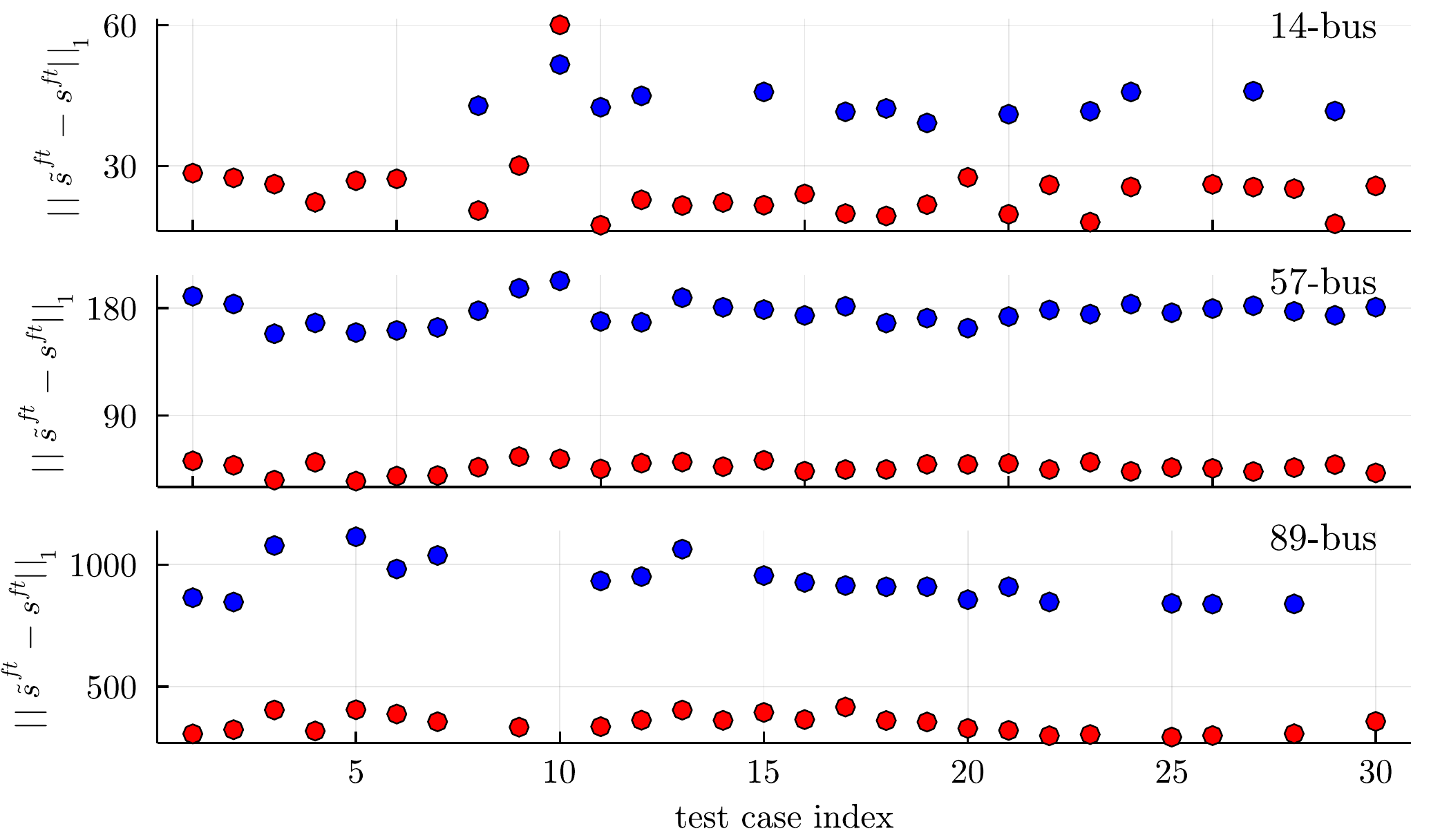}
    \caption{These three plots show the 1-norm error between the predicted apparent power flows ($\tilde{s}^{\text{ft}}$, calculated via the NN AC-UC and L AC-UC) and the actual apparent flows ($s^{\text{ft}}$, calculated via the MTP AC-OPF) in per-unit for the 30 test cases with varied loads. Red dots mark the 1-norm error associated with the NN-based power flow approximation, and blue dots mark the linear approximation. Missing data points correspond to infeasible solutions or problems that did not find a solution within a reasonable time.}
    \label{fig:Sf}
\end{figure}

Lastly, we compared the apparent power flows predicted in the NN-based and linear UC problems to those of the actual apparent power flows calculated via their associated MTP AC-OPFs.
Figure \ref{fig:Sf} compares the 1-norm error between the predicted and actual apparent power flows for the NN-based (red dots) and linear (blue dots) methods for the 30 considered loading cases in Table \ref{results_table}.
The NN-based approximation outperforms the linear approximation in all but one case. 
Apparent power flows in the opposite direction ($s^{\text{tf}}$) has similar results to those shown in Figure \ref{fig:Sf}.

\section{Conclusions}\label{sec:Conclusions}
This paper has demonstrated a proof-of-concept approach for modeling AC power flow constraints with a compact NN-based piecewise linear power flow mapping;
this mapping was learned directly from feasible power flow solutions. Once trained, we replaced the power flow constraints inside of the AC-UC problem with the NN-based model. 
Hence, the AC-UC MINLP is transformed into a more tractable MILP. 
Our results show that the NN-based formulation often generates feasible commitment schedules when the benchmark models (DC-UC and L AC-UC) could not. 
Furthermore, the NN-based formulation only produced infeasible schedules when both benchmark models did as well.
We also found, however, that the performance of the NN depended strongly on the power flow samples used to train the model. Future work will readily investigate better methods for collecting training data across more targeted regions as well as NN performance verification. 

\appendices
\section{}\label{App_Jacobian}
\small{
The power injection Jacobian $\bm{J}_{{pq}}\in\mathbb{R}^{2n\times2n}$ relates polar voltage ($v$, $\theta$) and nodal power injection ($p^{\rm inj}$, $q^{\rm inj}$) perturbations:
\begin{align}\label{eq: Jinj}
\bm{J}_{pq}=(\langle{\tt d}({\bm Y}_{b}{v}e^{j{\theta}})^{*}\rangle+\langle{\tt d}({v}e^{j{\theta}})\rangle N\langle {\bm Y}_{b}\rangle)R({v}e^{j{\theta}}),
\end{align}
where ${\tt d}(\cdot)$ is the diagonalization operator, and $R(\cdot)$, $N$, and $\langle \cdot \rangle$ are given in~\cite{Bolognani:2015}. The apparent power line flow Jacobians, $\bm{J}_{s,{\rm ft}},\bm{J}_{s,{\rm tf}},\in\mathbb{R}^{m\times2n}$, can be constructed by first partitioning the Jacobians relating active and reactive power flows in the lines:
\begin{subequations}
\begin{align}
\!\bm{J}_{\gamma} & =(\langle{\tt d}({\bm Y}_{\gamma}{v}e^{j{\theta}})^{*}{\bm E}_{i}\rangle\!+\!\langle{\tt d}({\bm E}_{i}{v}e^{j{\theta}})\rangle{N}\langle {\bm Y}_{\gamma}\rangle)\!R_{v}\\
 & =\left[\begin{array}{cc}
\frac{\partial{p^{\gamma}}}{\partial v} & \frac{\partial{p^{\gamma}}}{\partial{\theta}}\\
\frac{\partial{q^{\gamma}}}{\partial v} & \frac{\partial{q^{\gamma}}}{\partial{\theta}}
\end{array}\right],\; \gamma \in \{{\rm ft}, {\rm tf}\}, \; i \in \{1,2\}
\end{align}
\end{subequations}
where $R_{v}\triangleq R({v}e^{j{\theta}})$. When $\gamma={\rm ft}$, ${\bm E}_{i}={\bm E}_{1}$, and when $\gamma={\rm tf}$, ${\bm E}_{i}={\bm E}_{2}$, where ${\bm E}_{1} =\tfrac{1}{2}(|{\bm E}|+{\bm E})$, ${\bm E}_{2}=\tfrac{1}{2}(|{\bm E}|-{\bm E})$ are matrices which select the sending end and receiving end voltages, respectively. Since apparent power is related to active and reactive power via \eqref{S_fr_pq}-\eqref{S_to_pq}, the chain rule yields the Jacobian of $s$, where $x_{{\tt d}}\triangleq{\tt d}(x)$:
\begin{align}\label{eq: JSft}
\bm{J}_{s,\gamma} & =(s_{{\tt d}}^{\gamma})^{-1}\left[p_{{\tt d}}^{\gamma}\frac{\partial{p}^{\gamma}}{\partial{v}}+q_{{\tt d}}^{\gamma}\frac{\partial{q^{\gamma}}}{\partial v} \;\quad
p_{{\tt d}}^{\gamma}\frac{\partial p^{\gamma}}{\partial{\theta}}+q_{{\tt d}}^{\gamma}\frac{\partial{q}^{\gamma}}{\partial{\theta}}\right].
\end{align}

\bibliographystyle{IEEEtran}
\bibliography{references.bib}}

\begin{thebibliography}{10}
\providecommand{\url}[1]{#1}
\csname url@samestyle\endcsname
\providecommand{\newblock}{\relax}
\providecommand{\bibinfo}[2]{#2}
\providecommand{\BIBentrySTDinterwordspacing}{\spaceskip=0pt\relax}
\providecommand{\BIBentryALTinterwordstretchfactor}{4}
\providecommand{\BIBentryALTinterwordspacing}{\spaceskip=\fontdimen2\font plus
\BIBentryALTinterwordstretchfactor\fontdimen3\font minus
  \fontdimen4\font\relax}
\providecommand{\BIBforeignlanguage}[2]{{%
\expandafter\ifx\csname l@#1\endcsname\relax
\typeout{** WARNING: IEEEtran.bst: No hyphenation pattern has been}%
\typeout{** loaded for the language `#1'. Using the pattern for}%
\typeout{** the default language instead.}%
\else
\language=\csname l@#1\endcsname
\fi
#2}}
\providecommand{\BIBdecl}{\relax}
\BIBdecl

\bibitem{bienstock2015strong}
D.~Bienstock and A.~Verma, ``Strong {NP}-hardness of {AC} power flows
  feasibility,'' \emph{Operations Research Letters}, vol.~47, no.~6, pp.
  494--501, 2019.

\bibitem{Baker:2021}
K.~Baker, ``Solutions of {DC OPF} are never {AC} feasible,'' in
  \emph{Proceedings of the Twelfth ACM International Conference on Future
  Energy Systems}, 2021, pp. 264--268.

\bibitem{foster_dissertation}
J.~D. Foster, ``Mixed-integer quadratically-constrained programming,
  piecewise-linear approximation and error analysis with applications in power
  flow,'' dissertation, The University of Newcastle, Australia, School of
  Mathematical and Physical Sciences, November 2013.

\bibitem{Nanou:2021}
S.~I. Nanou, G.~N. Psarros, and S.~A. Papathanassiou, ``Network-constrained
  unit commitment with piecewise linear {AC} power flow constraints,''
  \emph{Electric Power Systems Research}, vol. 195, p. 107125, 2021.

\bibitem{donon2020neural}
B.~Donon, R.~Cl{\'e}ment, B.~Donnot, A.~Marot, I.~Guyon, and M.~Schoenauer,
  ``Neural networks for power flow: Graph neural solver,'' \emph{Electric Power
  Systems Research}, vol. 189, p. 106547, 2020.

\bibitem{Hu:2021}
X.~Hu, H.~Hu, S.~Verma, and Z.-L. Zhang, ``Physics-guided deep neural networks
  for power flow analysis,'' \emph{IEEE Transactions on Power Systems},
  vol.~36, no.~3, pp. 2082--2092, 2021.

\bibitem{tjeng2017evaluating}
V.~Tjeng, K.~Xiao, and R.~Tedrake, ``Evaluating robustness of neural networks
  with mixed integer programming,'' \emph{arXiv preprint arXiv:1711.07356},
  2017.

\bibitem{Venzke:2020}
I.~Murzakhanov, A.~{Venzke}, G.~S. {Misyris}, and S.~{Chatzivasileiadis},
  ``Neural networks for encoding dynamic security-constrained optimal power
  flow,'' \emph{arXiv e-prints}, p. arXiv:2003.07939, Oct. 2021.

\bibitem{venzke2020learning}
A.~Venzke, G.~Qu, S.~Low, and S.~Chatzivasileiadis, ``Learning optimal power
  flow: Worst-case guarantees for neural networks,'' in \emph{2020 IEEE
  International Conference on Communications, Control, and Computing
  Technologies for Smart Grids (SmartGridComm)}.\hskip 1em plus 0.5em minus
  0.4em\relax IEEE, 2020, pp. 1--7.

\bibitem{Katz:2020b}
J.~Katz, I.~Pappas, S.~Avraamidou, and E.~N. Pistikopoulos, ``The integration
  of explicit {MPC} and {ReLU} based neural networks,''
  \emph{IFAC-PapersOnLine}, vol.~53, no.~2, pp. 11\,350--11\,355, 2020.

\bibitem{Huchette:2020}
J.~Huchette, H.~Lu, H.~Esfandiari, and V.~Mirrokni, ``Contextual reserve price
  optimization in auctions via mixed-integer programming,'' \emph{arXiv
  preprint arXiv:2002.08841}, 2020.

\bibitem{Say:2017}
B.~Say, G.~Wu, Y.~Q. Zhou, and S.~Sanner, ``Nonlinear hybrid planning with deep
  net learned transition models and mixed-integer linear programming,'' in
  \emph{26th International Joint Conference on Artificial Intelligence
  (IJCAI)}, 2017, pp. 750--756.

\bibitem{Grimstad:2019}
B.~Grimstad and H.~Andersson, ``{ReLU} networks as surrogate models in
  mixed-integer linear programs,'' \emph{Computers \& Chemical Engineering},
  vol. 131, p. 106580, 2019.

\bibitem{Zhang:2020}
Y.~Zhang, C.~Chen, G.~Liu, T.~Hong, and F.~Qiu, ``Approximating trajectory
  constraints with machine learning--microgrid islanding with frequency
  constraints,'' \emph{IEEE Transactions on Power Systems}, vol.~36, no.~2, pp.
  1239--1249, 2020.

\bibitem{Zhang:2021}
Y.~Zhang, H.~Cui \emph{et~al.}, ``Encoding frequency constraints in preventive
  unit commitment using deep learning with region-of-interest active
  sampling,'' \emph{arXiv preprint arXiv:2102.09583}, 2021.

\bibitem{Cheng:2018}
Y.~Cheng, D.~Wang \emph{et~al.}, ``Model compression and acceleration for deep
  neural networks: The principles, progress, and challenges,'' \emph{IEEE
  Signal Processing Magazine}, vol.~35, no.~1, pp. 126--136, 2018.

\bibitem{Tseng:1996}
C.-L. Tseng, \emph{On power system generation unit commitment problems}.\hskip
  1em plus 0.5em minus 0.4em\relax University of California, Berkeley, 1996.

\bibitem{liu2018global}
J.~Liu, C.~D. Laird, J.~K. Scott, J.-P. Watson, and A.~Castillo, ``Global
  solution strategies for the network-constrained unit commitment problem with
  {AC} transmission constraints,'' \emph{IEEE Transactions on Power Systems},
  vol.~34, no.~2, pp. 1139--1150, 2018.

\bibitem{murillo-sanchez1998}
C.~Murillo-Sanchez and R.~J. Thomas, ``Thermal unit commitment including
  optimal {AC} power flow constraints,'' in \emph{31st Hawaii International
  Conference on System Sciences (HICSS)}, vol.~3, 1998, pp. 81--88.

\bibitem{castillo2016}
A.~Castillo, C.~Laird, C.~A. Silva-Monroy, J.-P. Watson, and R.~P. O’Neill,
  ``The unit commitment problem with {AC} optimal power flow constraints,''
  \emph{IEEE Transactions on Power Systems}, vol.~31, no.~6, pp. 4853--4866,
  2016.

\bibitem{zohrizadeh2018}
F.~Zohrizadeh, M.~Kheirandishfard \emph{et~al.}, ``Sequential relaxation of
  unit commitment with {AC} transmission constraints,'' in \emph{IEEE
  Conference on Decision and Control (CDC)}, December 2018, pp. 2408--2413.

\bibitem{donon2019graph}
B.~Donon, B.~Donnot, I.~Guyon, and A.~Marot, ``Graph neural solver for power
  systems,'' in \emph{2019 International Joint Conference on Neural Networks
  (IJCNN)}.\hskip 1em plus 0.5em minus 0.4em\relax IEEE, 2019, pp. 1--8.

\bibitem{Venzke:2021}
A.~Venzke and S.~Chatzivasileiadis, ``Verification of neural network behaviour:
  Formal guarantees for power system applications,'' \emph{IEEE Transactions on
  Smart Grid}, vol.~12, no.~1, pp. 383--397, 2021.

\bibitem{Coffrin:2018}
C.~Coffrin, R.~Bent, K.~Sundar, Y.~Ng, and M.~Lubin, ``Powermodels. jl: An
  open-source framework for exploring power flow formulations,'' in \emph{20th
  Power Systems Computation Conference (PSCC)}, 2018, pp. 1--8.

\bibitem{Shchetinin:2019}
D.~Shchetinin, T.~T. De~Rubira, and G.~Hug, ``Efficient bound tightening
  techniques for convex relaxations of {AC} optimal power flow,'' \emph{IEEE
  Transactions on Power Systems}, vol.~34, no.~5, pp. 3848--3857, 2019.

\bibitem{morales2012tight}
G.~Morales-Espa{\~n}a, J.~M. Latorre, and A.~Ramos, ``Tight and compact {MILP}
  formulation of start-up and shut-down ramping in unit commitment,''
  \emph{IEEE Transactions on Power Systems}, vol.~28, no.~2, pp. 1288--1296,
  2012.

\bibitem{Babaeinejadsarookolaee:2019}
S.~{Babaeinejadsarookolaee}, A.~{Birchfield} \emph{et~al.}, ``The power grid
  library for benchmarking {AC} optimal power flow algorithms,''
  \emph{arXiv:1908.02788}, Aug. 2019.

\bibitem{Xavier:2021}
\BIBentryALTinterwordspacing
A.~S. Xavier, A.~M. Kazachkov, and F.~Qiu, ``{ANL-CEEESA/UnitCommitment.jl:
  v0.2.2},'' Jul. 2021. [Online]. Available:
  \url{https://doi.org/10.5281/zenodo.5120043}
\BIBentrySTDinterwordspacing

\bibitem{Bolognani:2015}
S.~Bolognani and F.~Dörfler, ``Fast power system analysis via implicit
  linearization of the power flow manifold,'' in \emph{53rd Annual Allerton
  Conference on Communication, Control, and Computing (Allerton)}, 2015, pp.
  402--409.

\end{thebibliography}

\end{document}